\documentclass[lettersize,journal]{IEEEtran}
\usepackage{amsmath,amsfonts}
\usepackage{algorithmic}
\usepackage{algorithm}
\usepackage{array}
\usepackage[caption=false,font=normalsize,labelfont=sf,textfont=sf]{subfig}
\usepackage{textcomp}
\usepackage{stfloats}
\usepackage{url}
\usepackage{verbatim}
\usepackage{graphicx}
\usepackage{cite}
\usepackage{bbm} 				
\usepackage{multirow} 			
\usepackage{booktabs} 			
\usepackage{graphicx}
\usepackage{hyperref}
\usepackage{fancyhdr}

\makeatletter

\newcommand{\Rmnum}[1]{\expandafter\@slowromancap\romannumeral #1@}
\makeatother
\begin{document}
\title{PV-SSD: A Multi-Modal Point Cloud 3D Object Detector based on Projection Features and Voxel Features}

\author{Yongxin Shao, Aihong Tan, Zhetao Sun, Enhui Zheng, Tianhong Yan and Peng Liao
\thanks{Yongxin Shao, Aihong Tan, Zhetao Sun, Enhui Zheng, Tianhong Yan and Peng Liao are with the School of Mechanical and Electrical Engineering, China Jiliang Universty,Hanzhou,310018,China (email:syx1536505936@163.com;Tanah@cjlu.edu.cn; zhetaosun@163.com; ehzheng@cjlu.edu.cn; thyan@163.com; liaopeng202310@163.com). Aihong Tan is the Corresponding Author.
}}

\markboth{}%
{Shell \MakeLowercase{\textit{et al.}}: A Sample Article Using IEEEtran.cls for IEEE Journals}


\maketitle
\thispagestyle{fancy}
\renewcommand{\headrulewidth}{0pt}
\lhead{\centering{\small{This article has been accepted for publication in a future issue of this journal, but has not been fully edited. Content may change prior to final publication. Citation information: DOI 10.1109/TETCI.2024.3389710, IEEE Transactions on Emerging Topics in Computational Intelligence}}}
\lfoot{}
\cfoot{\small{Copyright \copyright 2024 IEEE. Personal use of this material is permitted. Permission from IEEE must be obtained for all other uses, in any current or future media, including reprinting/republishing this material for advertising or promotional purposes, creating new collective works, for resale or redistribution to servers or lists, or reuse of any copyrighted component of this work in other works.}}
\rfoot{}
\begin{abstract}
3D object detection using LiDAR is critical for autonomous driving. However, the point cloud data in autonomous driving scenarios is sparse. Converting the sparse point cloud into regular data representations (voxels or projection) often leads to information loss due to downsampling or excessive compression of feature information. This kind of information loss will adversely affect detection accuracy, especially for objects with fewer reflective points like cyclists. This paper proposes a multi-modal point cloud 3D object detector based on projection features and voxel features, which consists of two branches. One, called the voxel branch, is used to extract fine-grained local features. Another, called the projection branch, is used to extract projection features from a bird's-eye view and focus on the correlation of local features in the voxel branch. By feeding voxel features into the projection branch, we can compensate for the information loss in the projection branch while focusing on the correlation between neighboring local features in the voxel features. To achieve comprehensive feature fusion of voxel features and projection features, we propose a multi-modal feature fusion module (MSSFA). To further  mitigate the loss of crucial features caused by downsampling, we propose a voxel feature extraction method (VR-VFE), which samples feature points based on their importance for the detection task. To validate the effectiveness of our method, we tested it on the KITTI dataset and ONCE dataset. The experimental results show that our method has achieved significant improvement in the detection accuracy of objects with fewer reflection points like cyclists.
\end{abstract}


\section{Introduction}
\IEEEPARstart{T}{he} actual application of three-dimensional sensors \cite{sun2020disp,chen2020dsgn,chang2018pyramid,li2019stereo}, such as LiDAR, in autonomous driving scenarios has greatly enhanced the performance of autonomous driving environment perception systems. It not only can real-time detect moving objects in the current environment \cite{chen20153d,deng2017amodal,xu2017learning,gupta2014learning}, such as cars, pedestrians, and cyclists, accurately determining their positions and orientations, but also is not affected by lighting conditions or obstructions. However, the sparsity, disorder, and rotational invariance \cite{qi2017pointnet} of point cloud data pose challenges for directly applying established 2D object detection \cite{zhou2021tmfnet,barkur2022rscdnet,cong2022psnet,li2018power,8820084} techniques. Therefore, there are still some issues that need to be addressed in object detection methods based on 3D point cloud \cite{chen2021novel,pramanik2021granulated,wang2023centernet,zhou2021apnet}.

Currently, the point cloud-based 3D object detection methods can be mainly categorized into point-based \cite{shi2019pointrcnn,yang2019std} methods, projection-based methods \cite{simony2018complex,shao2023efficient,lang2019pointpillars,ali2018yolo3d}, and voxel-based methods \cite{zhou2018voxelnet,yan2018second}. The point-based methods typically utilize PointNet \cite{qi2017pointnet} or PointNet++ \cite{qi2017pointnet++} to extract features from raw point cloud data. In this sense, point-based methods can achieve better performance in terms of detection accuracy. Nonetheless, in autonomous driving scenarios, the point clouds generated by LiDAR exhibit relative sparsity. Point-based methods incur high computational costs, potentially squandering as much as 80\% of the total computation time when handling sparse data, rather than efficiently conducting feature extraction \cite{liu2019point}. Projection-based and voxel-based methods first encode local features within the same mesh (transforming unordered point clouds into regular voxels or projection) and then use convolutional neural networks for feature extraction. Both of these methods demand fewer computational resources than point-based methods but incur a certain level of information loss. For example, projection-based methods may lose the amount of point cloud features by over-compressing them in one direction during data processing \cite{deng2021multi}. While the voxel-based methods often result in the loss of fine-grained local features due to downsampling.
 
In autonomous driving scenarios, various objects (such as cars, pedestrians, and cyclists) need to be processed. For objects like cars with more reflection points, the feature point loss caused by projection-based or voxel-based methods would not have a significant impact on detection accuracy. However, smaller objects (such as cyclists and pedestrians) and objects located at a greater distance have fewer reflective points. Even a small loss of information can have a significant impact on detection accuracy. To mitigate the adverse effects of local information loss on the detection accuracy of objects with fewer reflective points, we proposea a multi-modal point cloud 3D object detector based on projection features and voxel features (PV-SSD).

We first designed a dual-branch feature extraction structure that includes the projection branch and the voxel branch. The voxel branch is used to extract local fine-grained features. The projection branch is used to extract point cloud projection features and consider the correlation between voxel local features. By feeding voxel features into the projection branch, we can compensate for the information loss in the projection branch while focusing on the correlation between neighboring local features in the voxel features. It is noteworthy that, to prevent misalignment between 3D voxel features and 2D projection features, the voxel features are converted into 2D features from bird's-eye view before being fed into the projection branch.

To integrate the point cloud features from two modalities (voxel features and projection features), we propose the Multi-modal Spatial-Semantic Feature Aggregation (MSSFA) module. The MSSFA module can comprehensively integrate voxel features and projection features that have the same spatial index.

In practical detection tasks, not all information contributes equally to the detection task. For objects such as cyclists, pedestrians, or distant objects, which have a lower number of reflective points, the loss of a few critical features may have a significant impact on detection accuracy. Therefore, preserving crucial local information is particularly important during the downsampling process. However, when using convolutional neural networks for voxel feature extraction, downsampling leads to the loss of some fine-grained local feature information. In order to tackle the aforementioned issues, we propose a new voxel feature extraction method called Variable Receptive Field Voxel Feature Extraction (VR-VFE) module. VR-VFE selects feature points for the detection task based on their weights (during the training process, the loss function provides implicit supervision, allowing the network to assign greater weights to the feature points that are advantageous for the detection task), aiming to retain critical feature points during downsampling. In comparison to existing state-of-the-art methods \cite{xu2022two,liu2022sms}, VR-VFE utilizes feature point weights for sampling, enabling the network to learn the importance of each feature point and thus retain a maximum number of relevant feature points. Subsequent experiments demonstrate the high accuracy achieved by our sampling method in detecting cyclists.

In summary, we make three-fold contributions:
\begin{itemize}
	\item{We propose a two-branch feature extraction network designed for extracting multi-modal features from point clouds.}
	\item{In the voxel branch, we propose a Voxel Feature Extraction method based on feature weight sampling (VR-VFE).}
	\item{We propose a multi-modal feature fusion module (MSSFA) that integrates voxel features and projection features.}
\end{itemize}

\section{Related Work}
\subsection{Point-based 3D object detection methods}
The point-based methods typically utilize PointNet \cite{qi2017pointnet} or PointNet++ \cite{qi2017pointnet++} to extract features from raw point cloud data. F-PointNet \cite{qi2018frustum} was the first method to perform 3D object detection directly using the raw point cloud. It uses image-based 2D detectors to generate region proposals, followed by further feature extraction and subsequent detection tasks on the point cloud within these proposals. Leveraging mature 2D detectors, F-PointNet achieves good recall, but its heavy reliance on 2D detectors often leads to poor performance under complex lighting conditions. Different from F-PointNet, Point RCNN \cite{shi2019pointrcnn} generates region proposals using point cloud data, thus avoiding the impact of 2D detectors on detection accuracy. Subsequently, STD \cite{yang2019std} proposed a method using spherical anchors for proposal generation, achieving high recall with reduced computation. Different from previous methods, Shi et al. \cite{shi2020pv} proposed a new paradigm for 3D object detection, PV-RCNN. It uses point-wise features extracted from the voxel feature extraction network to refine region proposals, striking a good balance between computational efficiency and accuracy.
\subsection{Projection-based 3D object detection method}
The projection-based methods typically project unordered point clouds into different views, transform them into regular pseudo-image representations, and complete subsequent feature extraction and detection tasks using 2D convolution. Based on the feature encoding method, projection-based methods can be classified into two categories. The first category, such as MV3D \cite{chen2017multi}, AVOD \cite{ku2018joint}, Complex YOLO \cite{simony2018complex}, and YOLO3D \cite{ali2018yolo3d}, project point clouds into height, density, and intensity maps for feature extraction. MV3D and AVOD employ multi-view fusion, projecting point clouds into bird's-eye and perspective views. Complex-YOLO and YOLO3D use single-view projection, projecting point clouds into a bird's-eye view. The second category converts point clouds into pillar structures. PointPillars \cite{lang2019pointpillars} first proposed this encoding method, which converts point clouds into a top-down pillar structure and utilizes this structure to perform feature encoding. H$^{2}$3D R-CNN \cite{deng2021multi}, following the same feature encoding method as PointPillars, reconstructs feature points from bird's-eye and perspective views into a 3D feature representation. This structure fully utilizes the complementary nature of bird's-eye and perspective views, achieving better detection accuracy. However, projecting point clouds into a specific view often leads to excessive compression of point cloud information, resulting in a significant loss of feature information.
\subsection{Voxel-based 3D object detection method}
To handle unstructured point clouds, VoxelNet \cite{zhou2018voxelnet}  first proposed a method that converts unordered point clouds into regular voxels and then extracts features using 3D convolutional neural networks. However, in most cases in autonomous driving scenarios, the majority of the generated voxels are filled with zeros. Extracting features using 3D convolution would result in a significant amount of ineffective computation. To address this issue, SECOND \cite{yan2018second} introduced the sparse convolution, which greatly improves computational efficiency. Voxel R-CNN \cite{deng2021voxel} extends SECOND into a two-stage method and proposes Voxel RoI pooling to aggregate 3D features for refining region proposals. This two-stage structure allows Voxel R-CNN to achieve good detection accuracy even in coarse-grained voxels. In subsequent work, it was found that in autonomous driving scenarios, it is possible to obtain not only the annotated bounding boxes and the foreground points they contain, but also the spatial distribution information of each foreground. Leveraging this information can greatly enhance the accuracy of boundary box prediction. In response to this finding, PartA$^{2}$ \cite{shi2020points} proposed a new two-stage architecture. It incorporates the prediction of the internal spatial distribution information of foreground objects in the first stage, thereby achieving high-quality 3D bounding box detection. Subsequently, CIA-SSD \cite{zheng2021cia} and SA-SSD \cite{he2020structure} address the problem of misalignment between bounding boxes and classification confidence. CIA-SSD uses an IoU-aware confidence correction module and a multi-task head module to improve the imbalance between classification confidence and regression confidence in 3D detection tasks. SA-SSD proposes an auxiliary network that can be turned off at any time to enhance the network's sensitivity to boundary information. In autonomous driving scenarios, the distribution of point clouds is sparse and uneven. This distribution poses challenges for point cloud sampling. To address this issue, SARPNET \cite{ye2020sarpnet} proposes a new low-level feature encoder and utilizes a shape attention mechanism to learn the 3D shape priors of objects. Voxel-based methods show outstanding performance in terms of accuracy and speed due to the superiority of the convolution structure, but the choice of voxel size is still sensitive. To address this problem, SMS-Net \cite{liu2022sms} proposes a sparse multi-scale voxel feature aggregation network, which divides the raw point cloud into voxels of different sizes to perceive 3D features in different perceptual fields. Although voxel-based methods can preserve more point cloud information compared to projection, voxel-based methods often suffer from the loss of fine-grained feature information due to downsampling.
\begin{figure}[!t]
	\centering
	\includegraphics[width=3.5in]{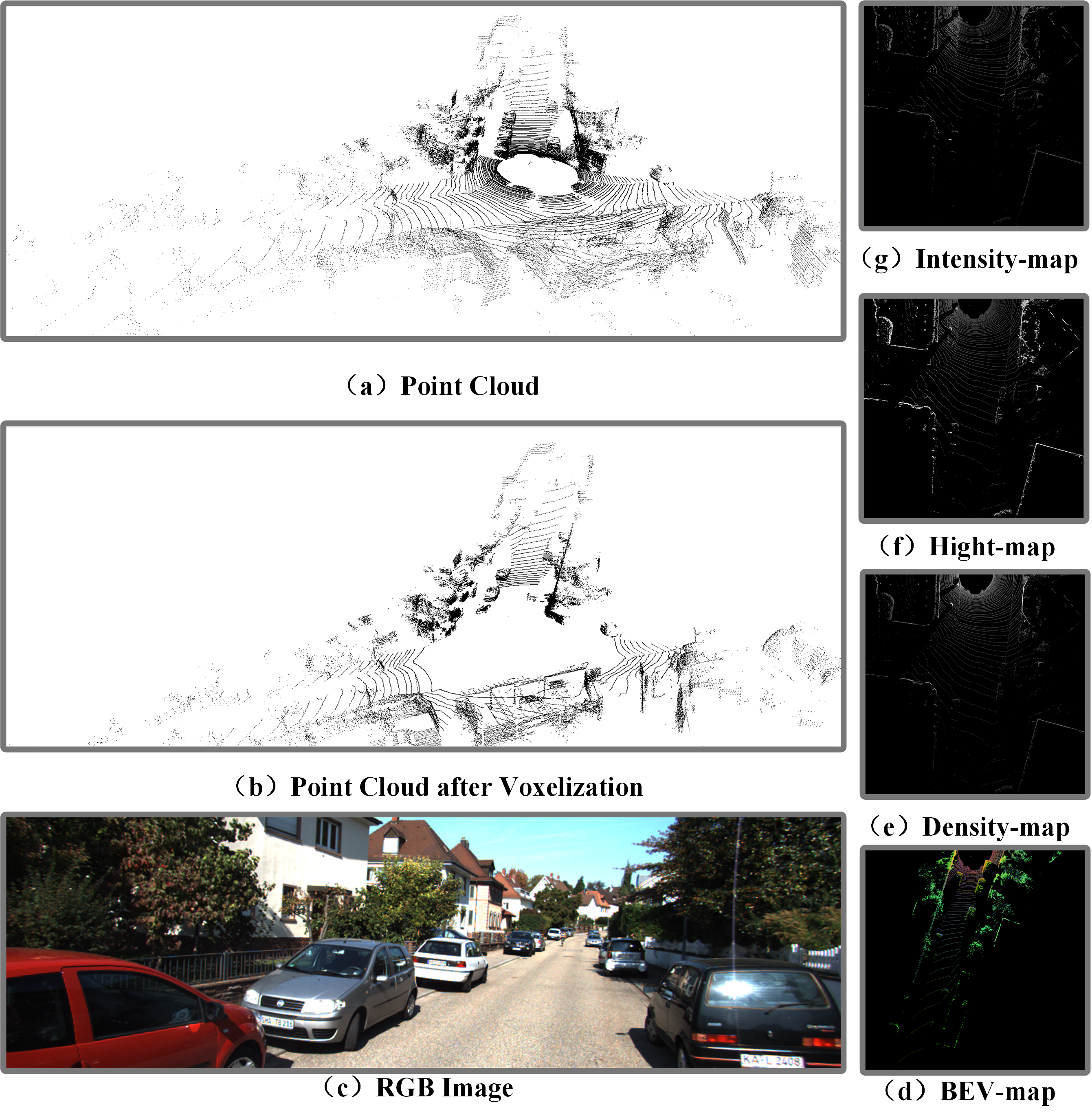}
	\caption{An illustration of (a) Point Cloud, (b) Point Cloud after voxelization, (c) RGB Image, (d) BEV-map, (e) Density-map, (f) Hight-map, and (g) Intensity-map. Each voxel in (b) contains 12 points, and the size of the voxel is $\left[0.1m,0.1m, 0.125m \right]$. Density-map, Height-map, and Intensity-map are the three channels of BEV-map. Each pixel in the density map, height map, and intensity map represents the normalized density of the point cloud, the maximum height of the point cloud, and the maximum intensity of the point cloud in that grid point, respectively.}
	\label{fig_1}
\end{figure}
\section{Methodology}
\subsection{Overview}
\begin{figure*}[!t]
	\centering
	\includegraphics[width=7in]{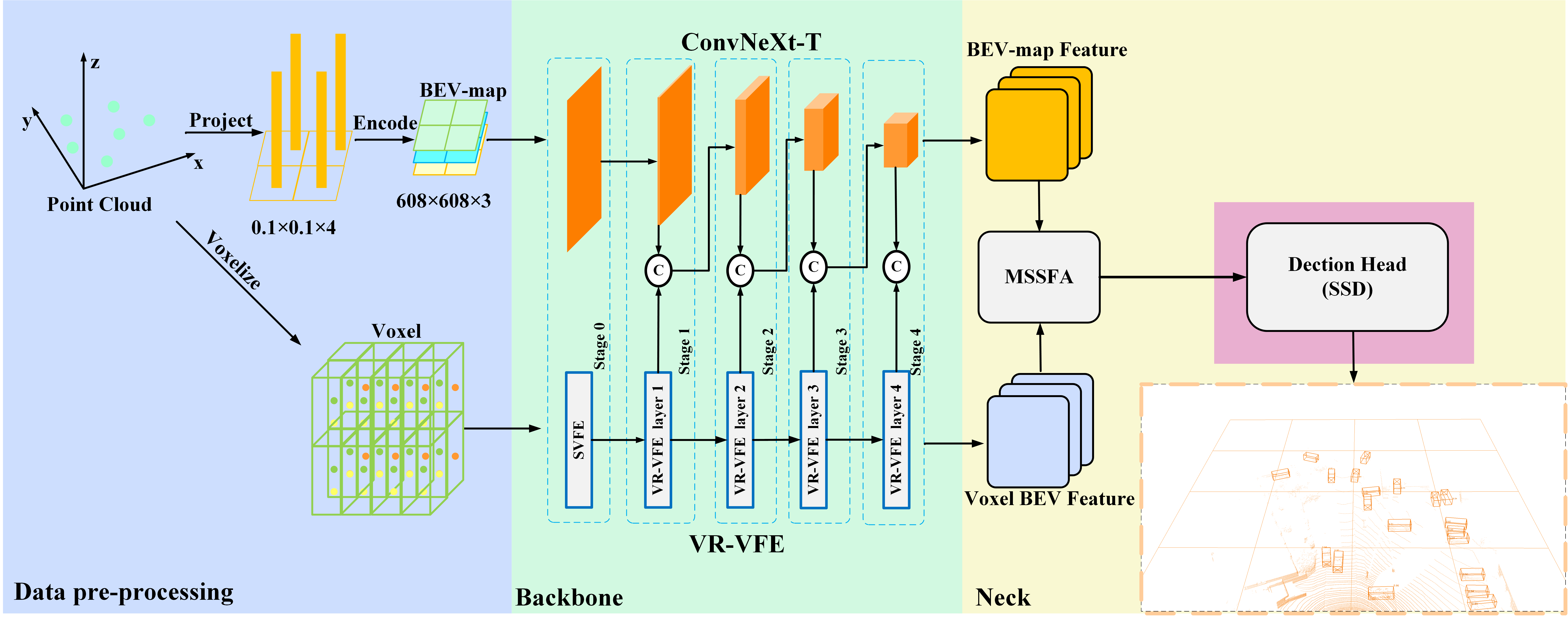}
	\caption{Overview of the PV-SSD structure. Where ‘C’ denotes concatenation operation (stitching by feature dimension). The point cloud data are transformed into BEV-map and voxel as input, the BEV-map feature is extracted with ConvNeXt-Tiny, and the voxel features are extracted with VR-VFE proposed in this paper. Multi-modal feature fusion is performed in MSSFA. Finally, the 3D detection is done in SSD Detection Head. ‘$0.1\times0.1\times4$’ refers to the size of each grid corresponding to reality after meshing the point cloud. ‘$608 \times 608 \times 3$’ refers to the size of the BEV-map generated by the projection.}
	\label{fig_2}
\end{figure*}

Fig. \hyperref[fig_1]{1} shows the form of the data covered in this paper. Fig. \hyperref[fig_1]{1} (d) shows the BEV-map, while Fig. \hyperref[fig_1]{1} (e)-(g) display its components; Fig. \hyperref[fig_1]{1} (a) shows the raw point cloud; Fig. \hyperref[fig_1]{1} (b) shows the Voxel; Fig. \hyperref[fig_1]{1} (c) shows the RGB Image. Our method consists of four parts, the flowchart is shown in Fig. \hyperref[fig_2]{2}.
\begin{itemize}
	\item{Data pre-processing: we use two different data pre-processing methods to obtain 2D projections of the bird's eye view (from now on, we call them BEV-map) and voxelized point cloud
		data.}
	\item{Backbone network: we feed the BEV-map into the 2D convolutional neural network and the voxelized point cloud data into the voxel feature extraction network to complete the feature extraction.}
	\item{Neck network: the BEV-map feature and the voxel features are fused with the features.}
	\item{Detection head: classify and regress 3D bounding boxes.}
\end{itemize}
\subsection{Data pre-processing}
Considering that the farther away from the LiDAR, the sparser the obtained point cloud. Therefore, we selected the point cloud within the range of $x \in \left [  -30.4m,30.4m \right ] $ and $y \in \left [  0m,60.8m \right ] $. Considering height of LiDAR above the ground and the problem of occlusion, we selected point cloud within the range of $z \in \left [  -3m,1m \right ] $. Here we define $P_{\Omega }$ as the set of point clouds, we adopt.
\begin{multline}
	\label{eq1}
	P_{\Omega }=\left \{ P=\left [ x,y,z \right ]^{T} \right \} \\ x \in \left[0m,60.8m\right],y \in \left[-30.4m,30.4m\right],z \in \left[-3m,1m\right]
\end{multline}

\textbf{Bev-map:} We adopt the same data pre-processing method as Complex-YOLO \cite{simony2018complex}, where the single-frame point cloud data in $P_{\Omega}$ is converted into pseudo-images in bird's-eye view. The maximum height, maximum intensity, and point cloud density in bird's-eye view will be encoded and filled into the corresponding channels. The final size of BEV-map is $608 \times 608 \times 3$, and each pixel corresponds to a realistic range of $\left[ 0.1m,0.1m,4m \right]$. The three channels of the BEV-map are encoded by the following \hyperref[eq2]{(2)}, \hyperref[eq3]{(3)}, and \hyperref[eq4]{(4)}.
\begin{equation}
	\label{eq2}
	z_{g}=max(P_{\Omega i \longrightarrow j} \cdot \left[0,0,1\right]^{T})
\end{equation}
\begin{equation}
	\label{eq3}
	z_{b}=max(I(P_{\Omega i \longrightarrow j}))
\end{equation}
\begin{equation}
	\label{eq4}
	z_{r}=min(1.0, log(N+1)/log(64)) \quad N=\left | P_{\Omega i \longrightarrow j} \right | 
\end{equation}

In the above \hyperref[eq2]{(2)}, \hyperref[eq3]{(3)}, and \hyperref[eq4]{(4)}, $z_{g}$ represents the maximum height; $z_{b}$ represents the maximum intensity; $z_{r}$ represents the normalized density; $I(P_{\Omega})$ represents the point cloud intensity; and $N$ represents the number of point clouds in each grid.

\textbf{Voxel}: We will voxelize the single-frame point cloud data in $P_{\Omega}$, using the same feature encoding as in PointPillars\cite{lang2019pointpillars}; each point cloud in the voxel will be encoded as a 10-dimensional vector $D$: $\left(x,y,z,r,x_{c},y_{c},z_{c},x_{p},y_{p},z_{p}\right)$.where $x,y,z,r$ represent the 3-dimensional coordinates of the point cloud and the reflection intensity, $x_{c},y_{c},z_{c}$ represent the geometric centers of all points in the voxel in which the point cloud is located, and $x_{p},y_{p},z_{p}$ are $x-x_{c},y-y_{c},z-z_{c}$, which represent the relative positions of the points to the geometric centers. Due to the sparsity of the point cloud, most of the sets of voxels are empty, while the non-empty voxels often have only a few points. This sparsity is exploited by imposing limits on the number of non-empty voxels per sample $(V)$ and the number of point clouds per voxel $(N)$ to create a dense tensor of size $(D, V, N)$. The data will be randomly sampled if a voxel has too many points to fit in this tensor. Conversely, if a voxel has too few points, it will be filled with zeros \cite{lang2019pointpillars}.
\begin{figure*}[!t]
	\centering
	\includegraphics[width=5in]{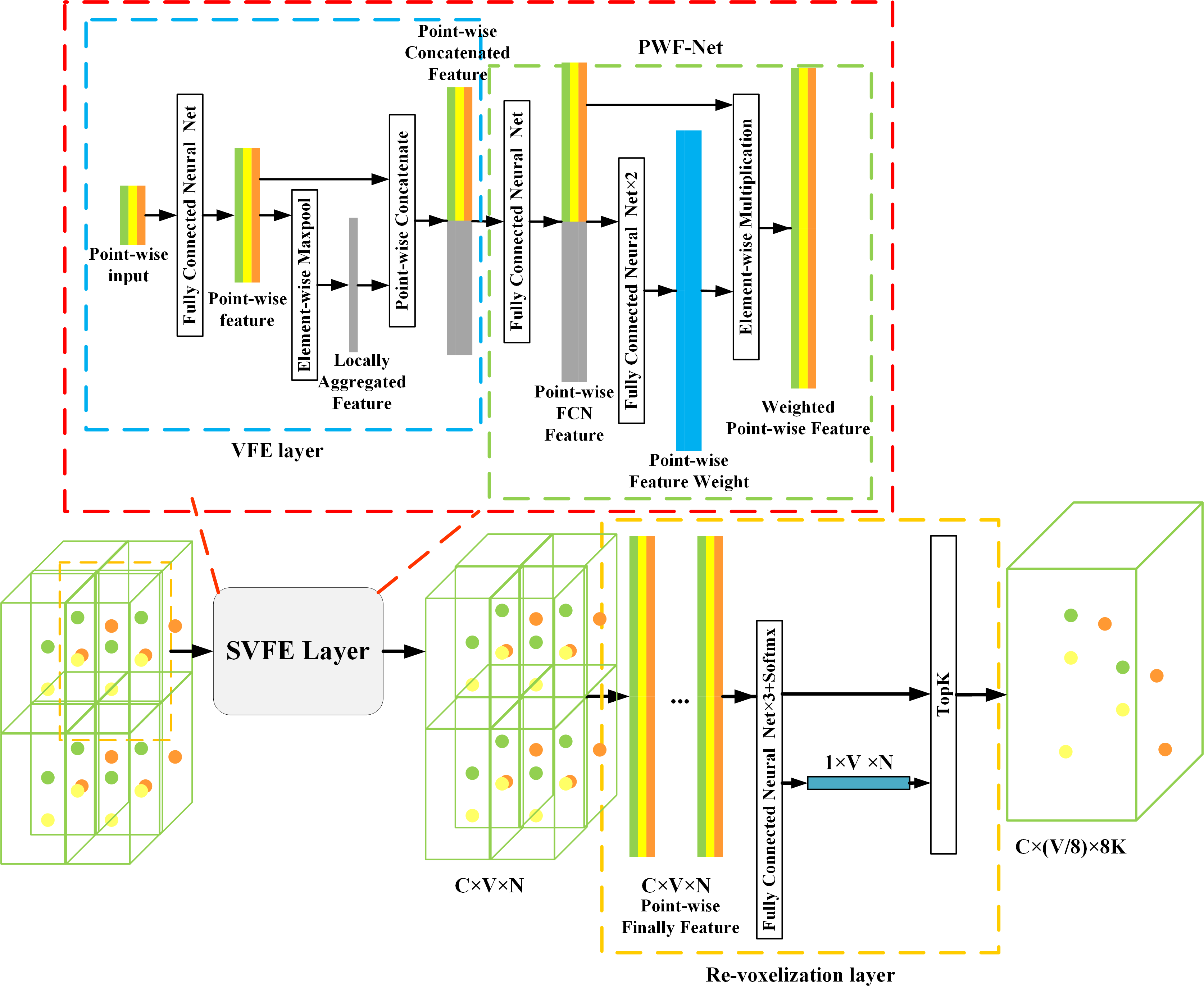}
	\caption{Overview of the VR-VFE structure. The red dashed box part is SVFE module; the blue dashed box part is VFE module; the green dashed box part is PFW-Net; the yellow dashed box part is Re-voxelization layer. In ‘$C\times V\times N$’, ‘$V$’ denotes the number of voxels, ‘$N$’ denotes that each voxel contains N point-wise, and ‘$C$’ denotes the number of features in each point-wise.}
	\label{fig_3}
\end{figure*}
\subsection{Backbone}
The backbone consists of BEV-map feature extraction network, voxel feature extraction network, and feature fusion.

\textbf{Projection Branch:} We use ConvNeXt \cite{liu2022convnet} as a feature extraction network for BEV-map, which is a feature extraction network proposed by Zhuang Liu et al. in 2022. ConvNeXt made a series of improvements to ResNet \cite{he2016deep} by borrowing some module designs from Transformer \cite{dosovitskiy2020image,liu2021swin} (e.g., replacing ReLU layer with GeLU layer, replacing Batch Normalize layer with Layer Normalize layer, etc.). In this paper, we use ConvNeXt-Tiny, a lighter version of ConvNext, for feature extraction of BEV-map.

\textbf{Voxel Branch:} The voxel feature extraction part is stacked by the VR-VFE layer. The VR-VFE layer consists of the modified Stacked Voxel Feature Encoding layer \cite{zhou2018voxelnet} (SVFE layer) and Re-voxelization layer. There are five stages in CovNeXt-Tiny, where Stage 0 is mainly used for downsampling to reduce the subsequent computational overhead, and Stages 1 to 4 are used for further feature extraction (each stage downsamples the feature maps by a factor of 2). Correspondingly, in the voxel branch, we also designed five stages, where the first stage consists of an SVFE layer for expanding the channel dimension of the voxel features (it should be noted that the first stage does not perform a downsampling operation). Each of the subsequent Stages 1 to 4 consists of a VR-VFE layer, which is used for further feature extraction (each of Stages 1 to 4 is downsampled, with 4-fold downsampling for Stage 1 and 2-fold downsampling for Stages 2 to 4, to avoid misalignment of features in the projection branch). The structure of VR-VFE layer is shown in Fig. \hyperref[fig_3]{3}. The structure of SVFE is shown in the red dashed box in Fig. \hyperref[fig_3]{3}. The VFE structure is shown in the blue dashed box in Fig. \hyperref[fig_3]{3}. The PFW-Net structure is shown in the green dashed box section of Fig. \hyperref[fig_3]{3}. The Re-voxelization layer structure is shown in the yellow dashed box section of Fig. \hyperref[fig_3]{3}. 

SVFE is a voxel feature encoding layer proposed in VoxelNet \cite{zhou2018voxelnet}. The SVFE module used in this paper replaces the original structure with a new one consisting of a VFE layer, a Fully Connected Network layer (FCN, each FCN consists of a fully connected layer, a GeLU layer, and a Layer Normalize layer), and a Point-wise Feature Weighting Net (PWF-Net). 

We use $VFE_{i}(c_{in},c_{out})$ to represent the VFE layer, where  $c_{in}$ represents the input point-wise feature dimension and $c_{out}$ represents the output feature dimension. Firstly, it raises the feature dimension to $c_{out}\setminus 2$ by FCN to get the point-wise feature. The point-wise feature is then operated by element-wise MaxPool to obtain locally aggregated feature, and finally it is concatenated with the point-wise feature by feature dimension to obtain point-wise concatenated feature with feature dimension $c_{out}$. 

During the voxel feature extraction process, we increase the channel dimension of feature points through fully connected layers to enable the extraction of richer semantic features in deep networks. However, as the channel dimension grows, features extracted in deeper layers of the network may contain redundant information that is irrelevant to the detection task. To address this issue, we introduce PFW-Net to highlight feature information that is more favorable for the detection task. PFW-Net can be seen as a simple self-attention mechanism that enables the network to focus more on features that are beneficial for object detection tasks. It computes the point-wise feature weight (between 0 and 1) using two fully connected layers and a Softmax layer, then multiplies it element-wise with the point-wise FCN feature to obtain the weighted point-wise feature. Additionally, during the network training phase, the loss function implicitly supervises network parameters, such as the weights and biases of fully connected layers and convolutional layers. This enables PFW-Net to derive a contribution value between 0 and 1 for each feature point similar to its importance to the final detection task. Subsequently, these contribution values are element-wise multiplied with the feature points to emphasize those that are more beneficial to the final detection task. In the subsequent ablation experiments, we further demonstrated the effectiveness of PWF-Net through empirical testing.

In the feature extraction process, we transform the voxelized point-wise features into Voxel-BEV features in the bird's-eye view for feature fusion with BEV-map features. However, each stage in the feature extraction of BEV-map involves a downsampling operation. For example, in ConvNeXt, stage 0 will do 4-fold downsampling for the feature map. The resolution corresponding to $608\times608$ in this paper will become $152 \times 152$, while the size of each grid point corresponding to the raw point cloud will become $\left[ 0.4m, 0.4m, 4m \right]$. At the same time, the SVFE layer will not change the size of each voxel. Therefore, we designed the Re-voxelization layer to ensure that the Voxel-BEV features at each stage have the same resolution as BEV-map features. On the other hand, as the network deepens, such operations can make the point-wise features in each voxel contain the richer semantic feature. 

For the detection task of objects with few reflection points, it is especially important to use a limited number of feature points to complete the feature extraction. In the sampling process, we want to be able to sample each point in the voxel according to its significance for the detection task, so we design a sampling method based on the weight of each point. We obtain the weight value of each point by changing the channels' number of the new point-wise feature to 1 through three fully connected layers and changing its value to between 0 and 1 by performing Softmax operation on the feature dimension. Finally, the top K points in each voxel are selected according to the weight value magnitude as the point-wise in each voxel. For example, if the original voxel contains 12 point-wise points and the voxel size is $\left[ 0.1m,0.1m,0.125m \right]$, we hope that after the Re-voxelization layer, each voxel contains 32 point-wise points. The first 4 points in each voxel will be selected according to the weight value (in this paper, we tend to select more points in the original voxel to increase the robustness of the network and then perform random sampling). After the sampling, to enhance the Point-wise spatial features in the resampled voxels, we encode the  $\left(x,y,z,r \right)$  of these points corresponding to the points in the raw point cloud with the same features as before to obtain $\left(x,y,z,r,x_{c},y_{c},z_{c},x_{p},y_{p},z_{p}\right)$, and stitch them with point-wise by feature dimension to obtain the new point-wise.
\begin{figure*}[!t]
	\centering
	\includegraphics[width=7in]{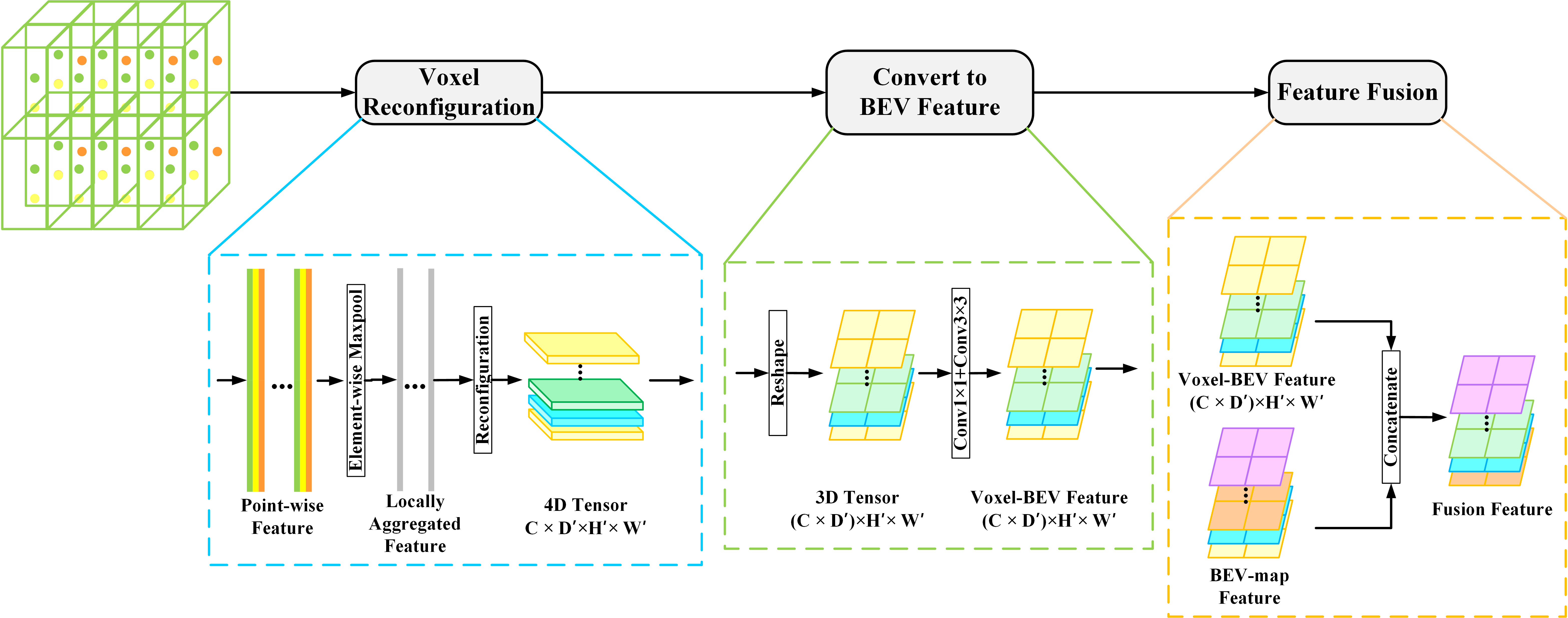}
	\caption{Feature fusion in the backbone. The blue dashed box shows the process of reconstructing the voxels into a 4D tensor. The green dashed box part transforms the reconstructed 4D tensor into a voxel feature map in the bird's eye view. The yellow dashed box part fuses the BEV-map features with the Voxel-BEV features for feature fusion.}
	\label{fig_4}
\end{figure*}

\textbf{Feature Fusion:} In the feature fusion, we first take the point-wise features (stored as $(C,V,N)$) in the voxels of each stage and get the local features (stored as $(C,1,N)$) that can represent each voxel through the element-wise MaxPool. The local voxel features are then reconstructed from $(C, 1, N)$ into a sparse 4D tensor of size $(C, D^{'}, H^{'}, W^{'})$, where $H^{'}$, $W^{'}$ represents the resolution of voxel features and $D^{'}$ represents the depth of voxel features. This 4D tensor is then reshaped into a 3D tensor of size $((C \times D^{'}), H^{'}, W^{'})$, and adjusted to a Voxel-BEV feature with a $1 \times 1$ convolution and a $3 \times 3$ convolution(the 3D voxel features will be transformed into 2D Voxel-BEV features by a $1 \times 1$ convolution and a $3 \times 3$ convolution to avoid the problem due to the misalignment of 3D voxel features with 2D BEV-map features), and finally, the BEV-map feature obtained from the same Stage as ConvNeXt is stitched by feature dimension. The feature fusion is completed by a $1 \times 1$ convolution and a $3 \times 3$ convolution, and the subsequent feature extraction is performed. It should be noted that we did not perform feature fusion on the features of Stage 0 considering that the main purpose of Stage 0 of ConvNeXt is to reduce computational consumption and not to perform too much feature extraction. In addition, if feature fusion is performed on the features of Stage 0, we need to create a tensor of $608 \times 608 \times C$ ($C$ is the number of channels) in the subsequent Multi-modal Spatial Feature Fusion (MSP-Fusion) module of MSSFA, which will bring a very large computational consumption. The specific structure is shown in Fig. \hyperref[fig_4]{4}. 
\begin{figure*}[!t]
	\centering
	\includegraphics[width=5in]{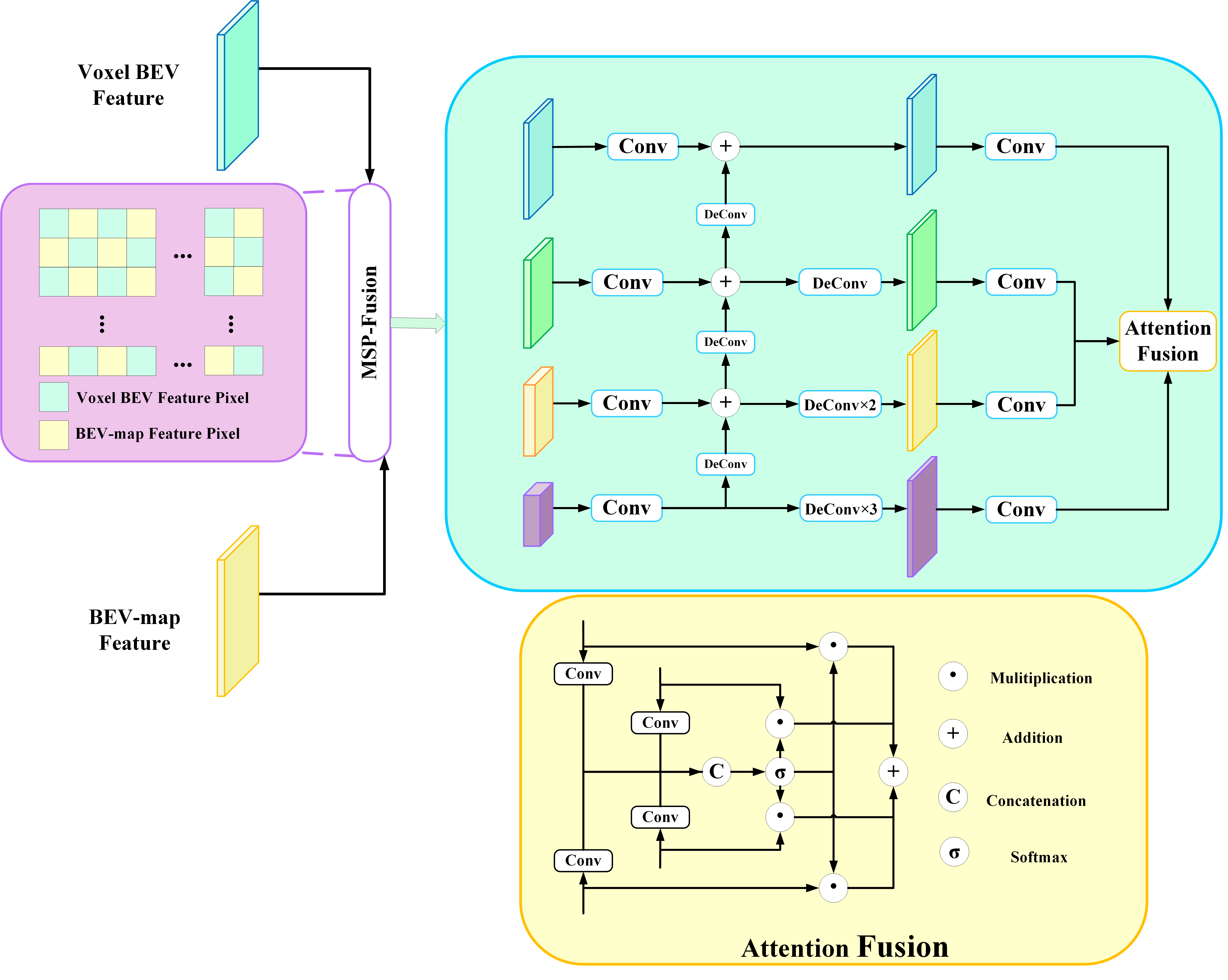}
	\caption{Overview of the MSSFA structure. The purple box is the feature fusion part of Voxel-BEV features and BEV-map features; the blue box is the adjustment part for adjusting the feature map size and feature number; the yellow box is the attention fusion module.}
	\label{fig_5}
\end{figure*}
\subsection{Neck}
In order to detect cars in autonomous driving scenarios, it is necessary to regress each car's precise location and classify the bounding box of each regression as positive or negative samples. In such a process, it is crucial to consider both low-level spatial features and high-level abstract semantic features. However, when we enrich the high-level abstract semantic features in the feature map by stacking convolutional layers, the resolution of the feature map gradually decreases, resulting in the loss of spatial information in the high-level feature map. In contrast, the low-level features retain more spatial information but contain less abstract semantic features. In CIA-SSD \cite{zheng2021cia}, the SSFA module is proposed for fusing feature information at different levels. In this paper, we improve the SSFA module by 1) propose the Multi-modal Spatial Feature Fusion module called MSP-Fusion; 2) using more features with different resolutions as input. We refer to the improved SSFA module as Multi-modal Spatial-Semantic Feature Aggregation module (MSSFA). The structure of MSSFA is shown in Fig. \hyperref[fig_5]{5}.

In the MSP-Fusion module, we perform feature fusion on two modal features with the same positional index and their neighboring features. The structure of MSP-Fusion is illustrated in the purple box in Fig. \hyperref[fig_5]{5}. Firstly, we transform the voxel feature into $((C \times D^{'}), H^{'}, W^{'})$ 3D tensor form. Then the number of features is adjusted to be the same as the BEV-map feature in this stage by a $1 \times 1$ convolution operation (with $C^{'}$ denoting the adjusted number of features). Then we will create a $((C^{'}, 2H^{'}, 2W^{'})$ tensor, perform adjacent interpolation operations on Voxel-BEV features with the same index and BEV-map features and complete the feature fusion in the Neck part by a $3\times3$ and a $1 \times 1$ convolution operations. The $3\times3$ convolution allows for the integration of multi-modal features in neighboring locations, while the $1 \times 1$ convolution allows for further separate feature integration for features at each location of the feature map. In this paper, we use four resolution features as inputs ($304 \times 304$, $152 \times 152$, $76 \times 76$, $38 \times 38$). To further demonstrate the effectiveness of MSP-Fusion, in the subsequent ablation experiments, we compare MSP-Fusion with two other fusion strategies: concatenating features of the two modalities along the feature dimension and directly adding features of the two modalities.
\subsection{Detection Head and Loss function}
In this paper, we use the same detection head as PointPillars, with single shot detector (SSD) \cite{liu2016ssd} settings for 3D object detection. We match the prior boxes to the ground truth using 2D Intersection over Union (IoU) \cite{everingham2010pascal}. Bounding box height and elevation are not used for matching; instead given a 2D match, the height and elevation become additional regression targets.

The loss function we use the same loss function as SECOND \cite{yan2018second} and PointPillars \cite{lang2019pointpillars}. Ground truth boxes and anchors are defined by $(x,y,z,w,l,h,\theta)$. The localization regression residuals between ground truth boxes and anchors are defined as \hyperref[eq5]{(5)}.
\begin{multline}
	\label{eq5}
	\bigtriangleup x = \frac{x^{gt}-x^{a}}{d^{a}}, \bigtriangleup y = \frac{y^{gt}-y^{a}}{d^{a}}, \bigtriangleup z = \frac{z^{gt}-z^{a}}{d^{a}} \\
	\bigtriangleup w = log \frac{w^{gt}}{w^{a}} ,\bigtriangleup l = log \frac{l^{gt}}{l^{a}} , \bigtriangleup h = log \frac{h^{gt}}{h^{a}}  
\end{multline}

Where $x^{gt}$, $x^{a}$  denote ground truth and anchor boxes, respectively; $d^{a}=\sqrt{(w^{a})^{2}+(l^{a})^{2}} $. The total localization loss can then be defined as:
\begin{equation}
	\label{eq6}
	L_{loc}=\sum_{b\in (x,y,z,w,h,l)} SmoothL1(\bigtriangleup b)
\end{equation}

We choose Focal loss \cite{lin2017focal} as the loss function for object classification. Where $\theta _{t}$  is the true value of the object orientation angle and $\theta _{gt}$ is the predicted value of the object orientation angle.
\begin{equation}
	\label{eq7}
	L_{cls}=-\theta _{t}log(\theta _{gt})-(1-\theta _{t})log(1-\theta _{gt})
\end{equation}

We choose Cross Entropy loss as the regression loss function for the object orientation angle.
\begin{equation}
	\label{eq8}
	L_{cls}=-\alpha_{a}(1-p^{\alpha})^{\gamma}log(p^{\alpha})
\end{equation}
Where $p^{\alpha}$ is the category probability of an anchor box. We use the parameter settings of $\alpha = 0.25$ and $\gamma=2$ from the original paper. Thus, the total loss is as shown in \hyperref[eq9]{(9)}. Where $N_{pos}$ represents the number of positive samples, while $\beta_{loc}=2$,$\beta_{cls}=1$,$\beta_{dir}=0.2$.
\begin{equation}
	\label{eq9}
	L=\frac{1}{N_{pos}}(\beta_{loc}L_{loc}+\beta_{dir}L_{dir}+\beta_{cls}L_{cls}) 
\end{equation}
\section{Experiments and Results}
\subsection{Datasets}
\begin{figure*}[!t]
	\centering
	\includegraphics[width=7in]{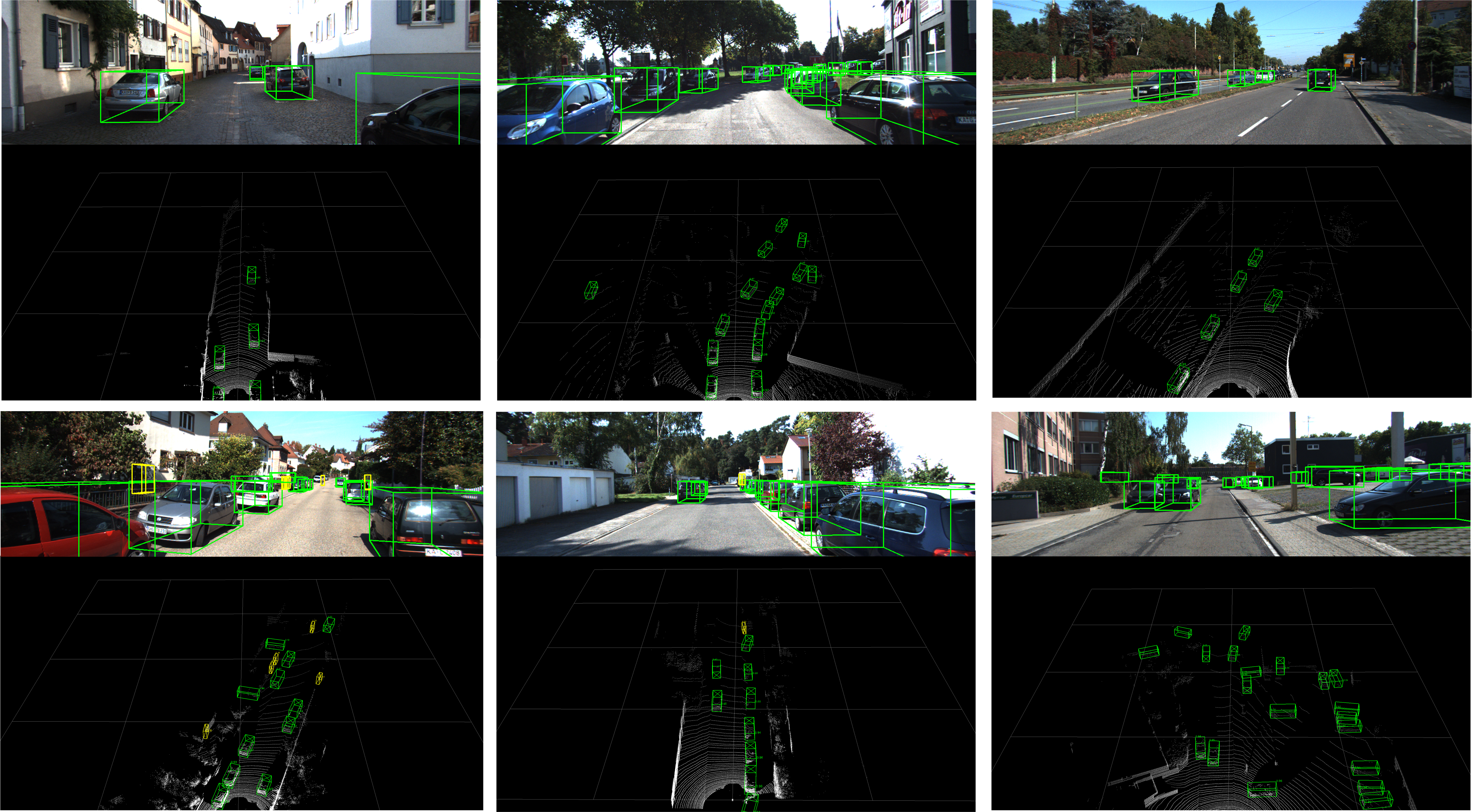}
	\caption{Qualitative analysis of KITTI results. The bottom half is the 3D bounding boxes on the point cloud. The upper part is the 3D bounding box projected into the image for a clearer view. Note that our method uses only LiDAR. We show the predicted boxes for ‘Car’ class (green) and ‘Cyclist’ class (yellow).}
	\label{fig_6}
\end{figure*}

We evaluate our method on the KITTI 3D object benchmark dataset \cite{geiger2012we}. The KITTI dataset has 7,481 training samples and 7,518 test samples. The training samples are divided into a training set (3,712 samples) and a validation set (3,769 samples). We conduct experiments on the most commonly used ‘Car’ and ‘Cyclist’ classes and evaluate the results by average precision (AP) and IoU thresholds (0.7 for cars and 0.5 for cyclists). Also, the dataset has three difficulty levels (easy, medium, and hard) based on object size, occlusion, and truncation levels. Fig. \hyperref[fig_6]{6} shows the results of our predictions.

\begin{figure*}[!t]
	\centering
	\includegraphics[width=7in]{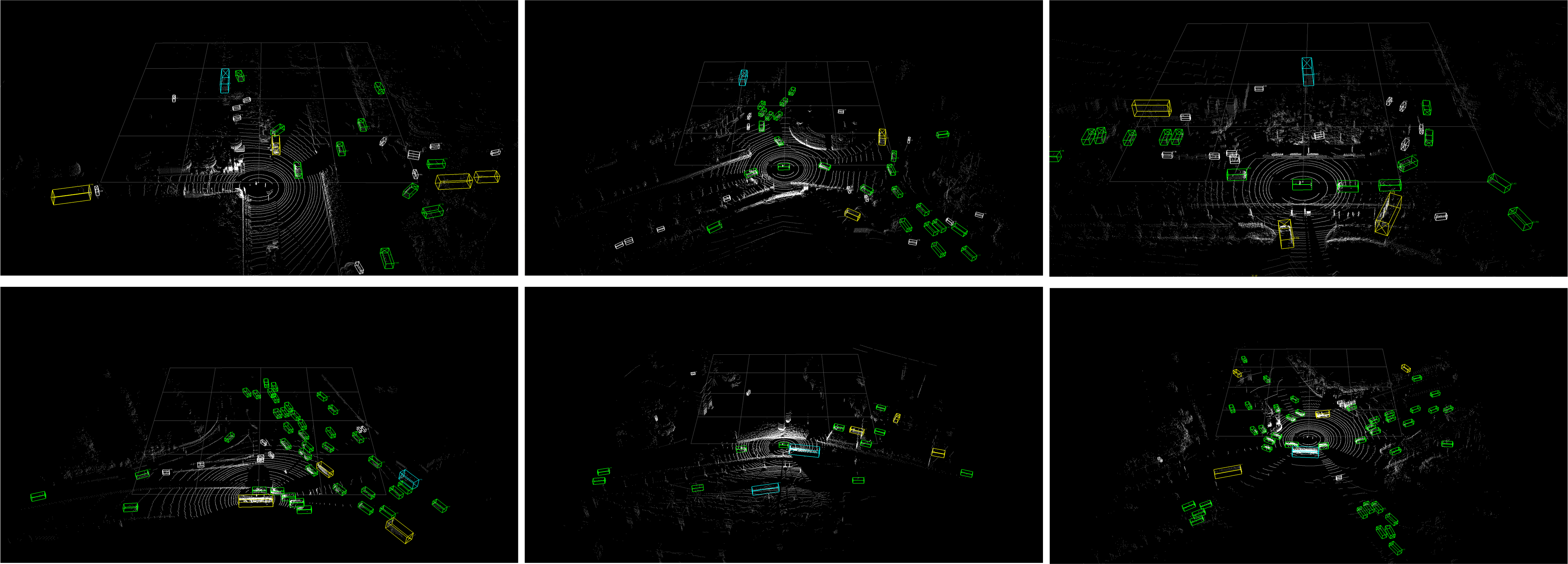}
	\caption{Qualitative analysis of ONCE results. Yellow boxes represent ‘Truck’, green boxes represent ‘Car’, blue boxes represent ‘Bus’, and white boxes represent ‘Cyclist’.}
	\label{fig_7}
\end{figure*}

To further validate the effectiveness of our method, we evaluate the performance of our method on the ONCE dataset \cite{mao2021one}. Six sequences are used for training, and four sequences are used for validation. We trained our method on the training set and evaluate it on the validation set. The ONCE dataset is divided into four evaluation levels based on the distance of objects (overall, objects within 0-30m range, objects within 30-50m range, objects beyond 50m range), and the evaluation metrics include Average Precision (AP) and Mean Average Precision (mAP). Fig. \hyperref[fig_7]{7} shows the results of our predictions.
\subsection{Implementation Details}
For KITTI dataset, we project the point cloud data in the range of $[0m,60.8m]$, $[-30.4m,30.4m]$. The point cloud data in the range of $[-3m,1m]$ (along the x-axis, y-axis, and z-axis) is projected into the bird's-eye view to construct a $608\times608$ BEV-map with each pixel corresponding to a realistic size of $[0.1m, 0.1m$]. The size of each voxel after voxelization is $[0.1m,0.1m,0.125m]$, and the number of point clouds is 12. For ONCE dataset, we project the point cloud data in the range of $[-60.8m,60.8m]$, $[-60.8m,60.8m]$. The point cloud data in the range of $[-5m,3m]$ (along the x-axis, y-axis, and z-axis) is projected into the bird's-eye view to construct a $608\times608$ BEV-map with each pixel corresponding to a realistic size of $[0.2m, 0.2m$]. The size of each voxel after voxelization is $[0.2m,0.2m,0.25m]$, and the number of point clouds is 12.
\begin{table*}[h]
	\label{table1}
	\caption {The 3D detection results on the KITTI test set for ‘Car’ and ‘Cyclist’ classes using AP@R40. "L" and "R" represent using LiDAR and RGB images, respectively. Note that we only compared with the 3D detection part of DIFI.}
	\centering
	\footnotesize	
	\resizebox{\linewidth}{!}{     				
		\begin{tabular}{cccccccccccccc}  	
			\toprule    				
			\multirow{2}{*}{Method} & 
			\multirow{2}{*}{Input} &    	
			\multicolumn{3}{c}{Car $AP_{Bev}$(\%)} & 
			\multicolumn{3}{c}{Cyclist $AP_{Bev}$(\%)} &
			\multicolumn{3}{c}{Car $AP_{3D}$(\%)} & 
			\multicolumn{3}{c}{Cyclist $AP_{3D}$(\%)} \\ 	  
			\cline{3-14}     			
			&  & Easy & Mod. & Hard & Easy & Mod. & Hard & Easy & Mod. & Hard & Easy & Mod. & Hard\\ 
			\midrule[0.5pt]
			MV3D\cite{chen2017multi} & L+I & 86.02&76.90 & 68.49& N/A& N/A&N/A &71.09& 62.35& 55.12& N/A& N/A& N/A\\
			Cont-Fuse\cite{liang2018deep} & L+I &88.81&85.83 &77.33 & N/A& N/A&N/A &82.54 &66.22&64.04 &N/A & N/A& N/A \\
			Roarnet\cite{shin2019roarnet} & L+I &88.20&79.41&70.02&N/A &N/A &N/A &83.71 &73.04 & 59.16&N/A & N/A&N/A\\
			DIFI\cite{10092267} & L+I &91.01&87.51&84.25&N/A &N/A &N/A &85.29 &76.59 & 71.75&N/A & N/A&N/A\\
			AVOD\cite{ku2018joint} & L+I &88.53&83.79&77.90&68.09&57.48&50.77&81.84&71.88&66.38&64.00&52.18&46.61\\
			F-PoinNet\cite{qi2018frustum} & L+I &88.70&84.00&75.33&75.38&61.96&54.68&81.20&70.39&62.19&71.69&56.77&50.9\\
			MSL3D\cite{chen2022msl3d} & L+I &N/A&N/A&N/A&N/A&N/A&N/A&87.27&81.15&76.56&76.74&62.27&56.20\\
			\midrule[0.5pt]
			SMS-Net\cite{liu2022sms} & L &N/A&N/A&N/A&N/A&N/A&N/A&87.01&76.21&70.45&75.35&60.23&53.37\\
			CIA-SSD\cite{zheng2021cia} & L &N/A&N/A&N/A&N/A&N/A&N/A&89.59&80.28&72.89&N/A&N/A&N/A\\
			Fast Point-RCNN\cite{chen2019fast} & L &90.87&87.84&80.52&68.09&57.48&50.77&85.29&77.40&70.24&N/A&N/A&N/A\\
			SA-SSD\cite{he2020structure} & L &95.03&91.03&85.96&75.38&61.96&54.68&88.75&79.79&74.16&N/A&N/A&N/A\\	
			Part-A$^{2}$\cite{shi2020points} & L &91.70&87.79&84.61&83.43&68.73&61.85&87.81&78.49&73.51&79.17&63.52&56.93\\
			PV-RCNN\cite{shi2020pv} & L &94.98&90.65&86.14&82.49&68.89&62.41&90.25&81.43&76.82&78.60&63.71&57.65\\
			PointRCNN\cite{shi2019pointrcnn} & L &92.13&87.39&82.72&82.56&67.24&60.28&86.96&75.64&70.70&74.96&58.82&52.53\\
			STD\cite{yang2019std} & L &94.74&89.19&86.42&81.36&67.23&59.35&87.95&79.91&75.09&78.69&61.59&55.30\\
			SECOND\cite{yan2018second} & L &88.07&79.37&77.95&73.67&56.04&48.78&83.13&73.66&66.20&70.51&53.85&46.90\\
			PointPillars\cite{lang2019pointpillars} & L &88.35&86.10&79.83&79.14&62.25&56.00&79.05&74.99&68.30&75.78&59.07&52.92\\
			H$^{2}$3D R-CNN\cite{deng2021multi} & L &82.85&88.87&86.07&82.76&67.90&60.49&90.43&81.55&77.22&78.67&62.74&55.78\\
			DVFENet\cite{he2021dvfenet} & L &90.93&87.68&84.60&82.29&67.40&60.71&86.20&79.18&74.58&78.73&62.00&55.18\\
			SARPNET\cite{ye2020sarpnet} & L &88.93&87.26&78.68&79.94&62.80&55.86&84.92&75.64&67.70&77.66&60.43&54.03\\
			Ours & L &91.32&87.46&84.38&81.58&66.93&60.21&86.71&78.53&74.09&78.84&63.00&55.85\\
			\bottomrule     			
	\end{tabular}}  
\end{table*}
\begin{table*}[h]
	\label{table2}
	\caption {The 3D detection results on the KITTI validation set for ‘Car’ and ‘Cyclist’ classes using AP@R40. ‘L’ and ‘R’ represent using LiDAR and RGB images, respectively.}
	\centering
	\footnotesize	
	\resizebox{\linewidth}{!}{     				
		\begin{tabular}{cccccccccccccc}  	
			\toprule    				
			\multirow{2}{*}{Method} & 
			\multirow{2}{*}{Modality} &    	
			\multicolumn{3}{c}{Car $AP_{Bev}$(\%)} & 
			\multicolumn{3}{c}{Cyclist $AP_{Bev}$(\%)} &
			\multicolumn{3}{c}{Car $AP_{3D}$(\%)} & 
			\multicolumn{3}{c}{Cyclist $AP_{3D}$(\%)} \\
			\cline{3-14}     		
			&  & Easy & Mod. & Hard & Easy & Mod. & Hard & Easy & Mod. & Hard & Easy & Mod. & Hard\\ 
			\midrule[0.5pt]
			Complex-YOLO\cite{simony2018complex} & LiDAR &85.89&77.40&77.33&72.37&63.36&60.27&67.72&64.00&63.01&68.17&58.32&54.30\\
			SECOND\cite{yan2018second}  & LiDAR &89.89&87.32&85.47&81.90&68.36&62.27&88.03&77.96&76.04&77.07&62.28&59.26\\
			Ours(VR-VFE) & LiDAR &89.77&87.80&83.62&83.02&67.20&66.88&87.61&76.43&72.44&80.00&66.20&62.15\\
			Ours & LiDAR &90.11&87.93&85.28&82.35&72.18&67.29&88.53&77.80&75.82&84.73&70.17&66.33\\
			\bottomrule   	
	\end{tabular}}  
\end{table*}
\begin{table*}[h]
	\label{table3}
	\caption {Results on the ONCE validation set.}
	\centering
	\resizebox{\linewidth}{!}{      				
		\begin{tabular}{cccccccccc}  	
			\toprule    				
			\multirow{2}{*}{Method} &  	
			\multicolumn{4}{c}{Vehicle} & 
			\multicolumn{4}{c}{Cyclist} &
			\multirow{2}{*}{mAP}\\ 	  
			\cline{2-9}   			
			&$overall$&$0-30m$&$30-50m$&$>50m$&$overall$&$0-30m$&$30-50m$&$>50m$&\\ 
			\midrule[0.5pt]
			PV-RCNN\cite{shi2020pv}&77.77&89.39&72.55&58.64&59.37&71.66&52.58&36.17&68.57\\
			Point RCNN\cite{shi2019pointrcnn}&52.09&74.45&40.89&16.81&29.84&46.03&20.94&5.46&40.97\\
			PointPillars\cite{lang2019pointpillars}&68.57&80.86&62.07&47.04&46.81&58.33&40.32&25.86&57.69\\
			CenterPoints\cite{yin2021center}&66.79&80.10&59.55&43.39&63.45&74.28&57.94&41.48&65.12\\
			SECOND\cite{yan2018second}&71.19&84.04&63.02&47.25&58.04&69.69&52.43&34.61&64.62\\
			Ours&69.24&84.14&64.29&42.67&54.06&69.32&44.89&25.18&61.65\\
			\bottomrule     
	\end{tabular}} 
\end{table*}
\begin{table}[h]
	\label{table4}
	\caption {Experimental results of comparison experiments with other methods. The detection speed is compared on the GTX1060.}
	\centering
	\resizebox{\linewidth}{!}{      				
		\begin{tabular}{c|cccccc} 	
			\toprule    				
			Method&PontRCNN\cite{shi2019pointrcnn}&PV-RCNN\cite{shi2020pv}&SECOND\cite{yan2018second}&PointPillars\cite{lang2019pointpillars}&PartA$^{2}$\cite{shi2020points}&Ours\\
			\midrule[0.5pt]
			FPS&3&3.8&11&15&3&5\\
			\bottomrule     			
	\end{tabular}} 
\end{table}

For the data augmentation part, we first creat a look-up table of ground truth 3D boxes for all categories and the associated point clouds that fall within these 3D boxes, following the approach of SECOND \cite{yan2018second}. Then, for each sample, we randomly select 15 ground truth samples for ‘Car’ class and 15 for ‘Cyclist’ class, respectively, and place them into the current point cloud. Next, these ground truth boxes are individually augmented with data. Each ground truth box is randomly rotated (drawn from $[-\pi/4, \pi/4]$) and translated (x, y, and z values are randomly drawn from $[0,0.5]$). Finally, two sets of global data augmentation are performed, a random mirror flip along the x-axis \cite{yang2018pixor} and a global scaling \cite{yan2018second,zhou2018voxelnet} (scaling is randomly drawn from $[0.95, 1.05]$). Please refer to OpenPCDet\footnote{https://github.com/open-mmlab/OpenPCDet} for the specific configuration. We implement the method of this paper using OpenPCDet and complete all the experiments. 

For the feature extraction network of BEV-map we have chosen ConvNeXt-Tiny \cite{liu2022convnet}. The voxel feature extraction network is stacked by VR-VFE layer. In the MSSFA module, we take the outputs of Stage1, Stage2, Stage3, and Stage4 of the feature voxel extraction network as its input (corresponding to resolutions of $152 \times 152$, $76 \times 76$, $38 \times 38$, $19 \times 19$). The BEV-map feature and Voxel-BEV feature of the same stage perform adjacent interpolation operations. Then the output is adjusted for the number of channels with a $1 \times 1$ convolution with 256 output channels, and the features are fused with a $3 \times 3$ convolution with stride of 1 and padding of 1. The 2D DeConv layer consists of a $3 \times 3$ convolution and a $3 \times 3$ deconvolution. In attention fusion, a $3 \times 3$ convolution is performed with an output channel of 1 to obtain the attention map. The head network settings obey the settings in PointPillars \cite{lang2019pointpillars}. 

We merge the validation set of the KITTI dataset with the training set to construct a new training set (3,712 samples in the original training set and 3,769 samples in the original validation set, with a total of 7,481 samples in the combined training set). The ADAM optimizer \cite{kingma2014adam} and OnecycleLR \cite{smith2019super} are applied to train 100 epochs with the initial learning rate set to 0.003, weight decay to 0.01, and momentum to 0.9.
\subsection{Compared with Others}
All detection results are measured using the official KITTI/ONCE evaluation detection metrics. For KITTI dataset, 2D detection is performed in the image plane; AOS evaluates the average direction of 2D detection (measured by BEV, IoU threshold of 0.7 for ‘Car’ class and 0.5 for ‘Cyclist’ class). For the ONCE dataset, the detection difficulty is divided according to the distance from the LiDAR (0-30m, 30-50m, overall). The experimental results are shown in Table \hyperref[table1]{\Rmnum{1}} ,Table \hyperref[table2]{\Rmnum{2}} ,and Table \hyperref[table3]{\Rmnum{3}}.

Table \hyperref[table1]{\Rmnum{1}} shows the 3D detection AP and BEV detection AP of our method on the KITTI test set. In both tables, we classify these methods into two categories based on the type of sensor used (methods such as MV3D use both camera and LiDAR, while methods such as PV-RCNN use LiDAR). From the table, we can see that our method’s AP for ‘Car’ class performs poorly, with 3D AP of 86.71\%, 78.53\%, and 74.09\% under Easy, Mod, and Hard conditions, respectively, which has a certain gap with advanced methods such as PV-RCNN. However, compared with projection-based 3D object detection methods\cite{ku2018joint,chen2017multi,lang2019pointpillars}, our method all have good advantages. For the ‘Cyclist’ class AP, our method has good performance. The 3D AP for ‘Cyclist’ class is 78.84\%, 63.00\%, and 55.85\%. 

As can be seen from Table \hyperref[table1]{\Rmnum{1}}, our method has 3\% lower 3D AP in the ‘Car’ class compared to the H$^{2}$3D R-CNN\cite{deng2021multi}, which is also based on the projection method. After our analysis, the reasons are: (1) The H$^{2}$3D R-CNN uses a dynamic voxelization feature coding method \cite{zhou2020end}, establishing a complete mapping relationship between points and voxels. In contrast, we use a hard voxelization approach \cite{zhou2018voxelnet}, which leads to uncertain voxel filling and information loss due to random sampling and discarding operations. (2) The projected features in the H$^{2}$3D R-CNN retain higher dimensional feature information through the MLP layer. In contrast, we use the same feature coding method as Complex-YOLO \cite{simony2018complex}, which over-compresses the features in the bird's-eye view, resulting in information loss. (3) H$^{2}$3D R-CNN is a two-stage 3D detector that fully uses the prior knowledge of the Region Proposal Network (RPN) module. In contrast, our method is a one-stage 3D detector, and the detection accuracy depends heavily on how well the features are extracted. However, we found that our method has good accuracy for objects with fewer reflection points like cyclists. We analyze the main reasons for this as: (1) However, the hard voxelization we use causes some information loss, we use a weight-based feature point sampling method in the Re-voxelization layer. For objects with fewer reflective points, such as cyclists, we can prioritize the feature points that are more beneficial to the detection task. (2) PWF-Net can filter out the point feature information more beneficial to the detection task. (3) For objects with few reflection points and small sizes, like the ‘Cyclist’ class, the voxel description in 3D space is more responsive to its features than the 2D perspective view. In the MSSFA module, we complemented the BEV-map features with Voxel-BEV features at the same grid points. In the subsequent ablation experiments, we can also see that the Re-voxelization layer, PWF-Net and MSP-Fusion module have a good improvement of the detection accuracy for the ‘Cyclist’ class. 
\begin{table*}[h]
	\label{table5}
	\caption {Performance comparison on the KITTI validation set, with AP calculated by AP@R40 for ‘Car’ and ‘Cyclist’ classes. ‘BEV’, ‘VOXEL’, ‘PWF-NET’, ‘MSSFA’, ‘RE-VOXELIZATION LAYER’, ‘RE-VOXELIZATION LAYER (RANDOMLY SAMPLE)’ denote BEV-MAP as input; voxel as input; PWF-NET module proposed in this paper; MSSFA module proposed in this paper; the Re-Voxelization Layer proposed in this paper; the Re-Voxelization Layer with randomly sampling instead of sampling by point weights. Bolded values are the best performances for all methods.}
	\centering
	\tiny	
	\resizebox{\linewidth}{!}
	{     				
		\begin{tabular}{ccccccc}  	
			\toprule    				
			\multicolumn{2}{c}{Improved Part} & Method A & Method B & Method C & Method D & Method E \\
			\hline   
			\multicolumn{2}{c}{BEV} &$\surd$ &&$\surd $&$\surd $&$\surd $\\
			\multicolumn{2}{c}{Voxel} &&$\surd $&$\surd $&$\surd $&$\surd $\\
			\multicolumn{2}{c}{PWF-Net} &&&&$\surd $&$\surd $\\
			\multicolumn{2}{c}{MSSFA} &$\surd $&$\surd $&$\surd $&$\surd $&$\surd $\\
			\multicolumn{2}{c}{Re-voxelization layer} &&&&&$\surd $\\
			\multicolumn{2}{c}{Re-voxelization layer(Randomly Sample)} &&$\surd $&$\surd $&$\surd $&\\
			\hline   
			Car AP&Easy&80.61&89.36&89.77&89.80&\textbf{90.11}\\
			BEV Detection&Mod.&66.83&87.00&87.80&87.52&\textbf{87.93}\\
			(IoU=0.7)&Hard&64.22&82.55&83.62&82.44&\textbf{85.28}\\
			\hline   
			Cyclist AP&Easy&54.46&78.03&83.02&83.18&\textbf{85.35}\\
			BEV Detection&Mod.&42.66&64.39&67.20&69.75&\textbf{72.18}\\
			(IoU=0.5)&Hard&41.44&60.41&66.88&65.76&\textbf{67.29}\\
			\hline   
			Car AP&Easy&62.15&85.76&87.61&87.45&\textbf{88.53}\\
			3D Detection&Mod.&52.11&75.96&76.43&76.88&\textbf{77.80}\\
			(IoU=0.7)&Hard&50.00&71.18&72.44&71.44&\textbf{75.82}\\
			\hline   
			Cyclist AP&Easy&44.00&76.37&80.00&82.02&\textbf{84.73}\\
			3D Detection&Mod.&42.12&59.63&66.20&67.65&\textbf{70.17}\\
			(IoU=0.5)&Hard&39.25&57.33&62.15&63.00&\textbf{66.33}\\
			\bottomrule   	
	\end{tabular}}  
\end{table*}
\begin{table}[h]
	\label{table6}
	\caption {Analysis of the multi-features fusion operation. Notation: ‘MSP’ represents MSP-Fusion module. Bolded values are the best performances for all methods.}
	\centering
	\resizebox{\linewidth}{!}{      				
		\begin{tabular}{ccccccc} 	
			\toprule    				
			\multirow{2}{*}{Method} & \multicolumn{4}{c}{Feature Fusion} & \multicolumn{2}{c}{Mod. AP$_{3D}(\%)$} \\
			\cline{2-7}
			&BEV&Voxel&MSP&Concat&Car&Cyclist\\
			\midrule[0.5pt]
			A&$\surd $&&&&52.11&42.12\\
			B&&$\surd $&&&75.96&59.63\\
			F&$\surd $&$\surd $&&$\surd $&75.49&62.29\\
			C&$\surd $&$\surd $&$\surd $&&76.43&66.20\\
			\bottomrule     			
	\end{tabular}} 
\end{table}

Considering that the detection results for the KITTI test set are not submitted in Complex-YOLO, we compare our method with Complex-YOLO \cite{simony2018complex} on the val set of KITTI. In this paper, the same bird's-eye view projection encoding method as Complex-YOLO is used. As can be seen in Table \hyperref[table2]{\Rmnum{2}}, our method is more advantageous in both BEV AP and 3D AP. Although in 3D detection, Complex-YOLO does not predict the height of the object, which affects the accuracy of 3D detection to some extent, our method has a greater improvement compared with it in the BEV detection results. In addition, in Table \hyperref[table2]{\Rmnum{2}}, we compare VR-VFE with the feature extraction network using sparse convolution \cite{yan2018second} (referred to as SECOND in Table \hyperref[table2]{\Rmnum{2}}). From Table \hyperref[table2]{\Rmnum{2}}), it can be observed that, compared to SECOND, using VR-VFE as the feature extraction network shows a significant advantage in the accuracy of ‘Cyclist’. This is attributed to the weight-sampling method employed in VR-VFE during feature point downsampling. For objects with fewer reflected points, such as cyclists, the weight-sampling method helps to preserve crucial feature points to a large extent during the downsampling process.

Table \hyperref[table3]{\Rmnum{3}} shows the detection results of the method proposed in this paper on the validation set of the ONCE dataset. From Table \hyperref[table3]{\Rmnum{3}}, it can be seen that the proposed method performs well in large-scale complex LiDAR scenes. Compared to the projection-based method PointPillars, our method demonstrates a good precision advantage on the ONCE dataset. In Fig. \hyperref[fig_7]{7}, we present the visualized detection results on the ONCE dataset.

To validate the detection speed of our method, we compare it with several other methods on a device equipped with GTX1060. As shown in Table \hyperref[table4]{\Rmnum{4}}, it is evident that due to the presence of a multi-branch structure, our method does not perform well in terms of detection speed.
\subsection{Ablation Experiment}
To further validate the validate the effectiveness of each part of our method, we performed further validation on the validation set of the KITTI dataset. All experiments are per-formed on the same data set division (3,712 samples for the training set and 3,769 samples for the validation set), using the same parameter configuration and 100 epochs trained. The experimental results are shown in Table \hyperref[table5]{\Rmnum{5}}.

\textbf{Method A:} Method A uses BEV-map as input and ConvNeXt-Tiny for feature extraction. Because only BEV-map is used as is input, the feature fusion part is removed from the MSSFA module that is not available in Method A. The resolution of BEV-map is 608×608, and the used point cloud range is $[60.8m, 60.8m, 4m]$. The rest of the settings are the same as the method proposed in this paper. As can be seen from Table  \hyperref[table5]{\Rmnum{5}}, the detection results obtained by using only BEV-map as input are poor. This is because projecting the point cloud into the bird’s view causes more information loss in the data processing stage, which is not beneficial for the subsequent detection tasks.

\textbf{Method B:} Method B uses voxel as input, and a voxel feature extraction network constructed by stacking SVFE and Re-voxelization (Randomly Sample) layer is used for feature extraction. Because only voxel is used as input, the feature fusion part is removed from the MSSFA module that is not available in Method B. The used point cloud range is $[60.8m,60.8m,4m]$. The size of each voxel is $[0.1m,0.1m,0.125m]$, and each voxel is randomly sampled with 12 points. The rest of the settings are the same as the method proposed in this paper. Compared with Method A, the voxel can retain more point cloud information, and the Re-voxelization layer is designed to achieve feature interaction of adjacent voxels. Compared with Method A, the detection accuracy of Method B is greatly improved.

\textbf{Method C:} Method C uses a mixture of BEV-map and voxel as input. Compared with methods A and B, the fusion of voxel and BEV-map features is added to the MSSFA module. BEV-map and voxel have the same parameters as methods A and B. The rest of the settings are the same as the method proposed in this paper. The detection accuracy has been improved compared to Method A and Method B (compared to Method A, 3D AP of ‘Car’ class is improved by 25.46\%, 24.32\%, 22.44\%, 3D AP of ‘Cyclist’ class is improved by 36\%, 22.08\%, 22.9\%; compared to Method B, 3D AP of ‘Car’ class is improved by 1.85\%, 0.47\%, 0.56\%, 3D AP of ‘Cyclist’ class is improved by 3.63\%, 6.57\%, and 4.82\%).

\textbf{Method D:} Method D adds PWF-Net to Method C. The rest of the settings are the same as the method proposed in this paper. Compared to Method C, Method D has improved detection accuracy in ‘Cyclist’ class (compared to Method C, 3D AP of ‘Car’ class is improved by 2.02\%, 1.35\%, 0.85\%, 3D AP of ‘Cyclist’ class is improved by 2.02\%, 1.35\%, 0.85\%).

\textbf{Method E:} Method E (the method proposed in this paper) replaces random sampling with feature weight-based sampling. Compared with Method D, the detection accuracy of Method E is improved more in the detection accuracy of small objects like cyclist with fewer point clouds (compared with Method D, the 3D AP of ‘Car’ is improved by 1.08\%, 0.92\%, 4.38\%, the 3D AP of ‘Cyclist’ class is improved by 4.71\%,2.52\%, 3.33\%).

By comparing Method A and C, we can find that voxel can be used to compensate for the information loss of the point cloud during projection. Compared with Method B, Method C adds point cloud projection as input, which can also help the final detection task. The comparison between Method C, D, and E shows that the PWF-Net and the Re-voxelization layer sampled by weight are good for detecting objects with few reflection points like cyclist. This is because, for objects with fewer points, the Re-voxelization layer with weight sampling and PWF-Net can filter out the feature points that are more beneficial for the detection task.

In the MSSFA module, we add the feature fusion module for Voxel-BEV and BEV-map feature fusion. To verify the effective ness of our feature fusion module design, we design the corresponding comparison experiments. Table \hyperref[table6]{\Rmnum{6}} shows the experimental results. Method A, B, and C, in Table  \hyperref[table6]{\Rmnum{6}}, corresponding to the methods mentioned in Table \hyperref[table5]{\Rmnum{5}}. We modify the feature fusion module in MSSFA to Concat operation (i.e., stitching by feature dimension) in Method F. The rest of the module design in Method F is the same as Method C. As we can see from Table \hyperref[table6]{\Rmnum{6}}, compared to Method F, Method C has improved 3D AP in Mod. conditions (‘Car’:0.94\%, ‘Cyclist’:3.91\%). The 3D AP of ‘Car’ and ‘Cyclist’ are also improved by Method F under Mod. conditions compared to Method A, B (‘Car’: 23\% compared to Method A, B, ‘Cyclist’: 20.17\%, 2.66\%; ‘Car’ by 23.38\%, -0.47\%, respectively).

\section{Conclusion}
In this paper, we propose a multi-modal point cloud 3D object detector based on projection features and voxel features. To address the information loss caused by point cloud projections or downsampling, we design a dual-branch network structure. The voxel branch is used to extract local fine-grained features, while the projection branch is used to extract point cloud projection features and consider the correlation between voxel local features. By feeding the voxel features into the projection branch, we can compensate for the information loss while focusing on the correlation between adjacent local features in the voxel features. Compared to previous projection-based methods such as Complex YOLO, PointPillars, MV3D, we have achieved significant improvements in the detection accuracy for both cars and cyclists. In addition, through ablation experiments, we compare the detection results of three different structures: single projection branch, single voxel branch, and voxel-projection dual branch. We found that the dual branch structure outperforms the single branch structure, especially for objects with fewer reflection points like cyclists (achieving a 10\% improvement compared to the single projection branch and a 5\%-7\% improvement compared to the single voxel branch). This fully demonstrates that the proposed dual branch structure can reduce the loss caused by projection.

Furthermore, when extracting voxel features using convolutional neural networks, there is a loss of fine-grained crucial information due to downsampling. This information loss has a significant impact on objects with fewer reflection points, such as cyclists. We propose a sampling method based on feature point weights. Subsequent experiments have also found that although there is a gap in detection accuracy for cars compared to the latest work H$^{2}$3D R-CNN, our method has a good competitiveness in detecting cyclists. In addition, in the ablation experiments, we compare the detection results using random sampling and feature point weight-based sampling. It is found that the method based on feature point weight sampling achieved a 2\%-3\% improvement.

However, it should be noted that while the design of the dual branch structure brings certain improvements in detection accuracy to our method, it also introduces a certain computational cost. Therefore, in terms of detection speed, our method does not have a significant advantage. In further research, considering the high computational cost of ConvNext, it is possible to try using a lighter network to some extent to improve the detection speed.
\section*{Acknowledgments}
This research was funded by the Natural Science Foundation of Zhejiang Province, grant number LY21F010013.
\bibliographystyle{IEEEtran}
\bibliography{reference.bib}


\end{document}